\documentclass[sigplan,screen, nonacm]{acmart}
\AtBeginDocument{%
  }

\setcopyright{acmlicensed}
\copyrightyear{2026}
\acmYear{2026}
\acmDOI{XXXXXXX.XXXXXXX}
\acmConference[ACM AI Leadership Summit '26, Breakthrough Impact Highlights Track]{Breakthrough Impact Highlights Track}{August 30--September 02, 2026}{Atlanta, GA, USA}
\usepackage{subfig}

\usepackage{wasysym}
\usepackage{xcolor}
\newcommand{\usenixhref}[3][black]{\href{#2}{\color{#1}{#3}}}

\definecolor{LimeGreen}{RGB}{51, 205, 51}
\usepackage[misc]{ifsym}

\newcommand{\mkletter}[0]{{\color{LimeGreen}{\Letter}}}
\definecolor{clpanw}{RGB}{85, 43, 111}
\definecolor{cluci}{RGB}{219, 109, 0}
\definecolor{clpurdue}{RGB}{226, 32, 17}
\definecolor{clkhalifa}{RGB}{200, 132, 157}
\definecolor{clleeds}{RGB}{100, 32, 57}
\definecolor{clyale}{RGB}{10, 10, 137}
\definecolor{clutdallas}{RGB}{75, 102, 131}

\newcommand{\mkatt}[0]{{\color{clpanw}{$^\ddag$}}}
\newcommand{\mkgsma}[0]{{\color{cluci}{$^\Phi$}}}
\newcommand{\mkrelai}[0]{{\color{clpurdue}{$^\S$}}}
\newcommand{\mkkhalifa}[0]{{\color{clkhalifa}{$^\Upsilon$}}}
\newcommand{\mkkleeds}[0]{{\color{clleeds}{$^\Psi$}}}
\newcommand{\mkkyale}[0]{{\color{clyale}{$^\Omega$}}}
\newcommand{\mkutdallas}[0]{{\color{clutdallas}{$^\P$}}}
\usepackage{eso-pic}
\usepackage{xcolor}

\AddToShipoutPictureFG*{%
  \AtPageUpperLeft{%
    \hspace{0.5\paperwidth}%
    \raisebox{-0.45in}{%
      \makebox[0pt][c]{%
        \textcolor{blue}{\textit{Accepted at the ACM AI Leadership Summit, Breakthrough Impact Highlights Track, 2026.}}%
      }%
    }%
  }%
}

\begin{document}

\title[OTel]{\texttt{OTel}: Building Domain-Specialized Telecom LLM Foundations for Intelligent Networks}


\author{
    % Anonymous submission \# 88 
    \usenixhref{}{\rm Farbod Tavakkoli}\mkatt \mkletter,
    \usenixhref{}{\rm Roderic Paulk}\mkatt,
    \usenixhref{}{\rm Jorden Terrazas}\mkatt,
    \usenixhref{}{\rm Kenneth Church}\mkatt,
    \usenixhref{}{\rm Mark Austin}\mkatt,
    \usenixhref{}{\rm Louis Powell}\mkgsma \mkletter,
    \usenixhref{}{\rm Gregory Diamos}\mkrelai,
    \usenixhref{}{\rm Lina Bariah}\mkkhalifa,
    \usenixhref{}{\rm Syed Ali Raza Zaidi}\mkkleeds,
    \usenixhref{}{\rm Maryam Hafeez}\mkkleeds,
    \usenixhref{}{\rm Ali Maatouk}\mkkyale,
    \usenixhref{}{\rm Imtiaz Karim}\mkutdallas \mkletter
    \\
    \mkatt \usenixhref{https://about.att.com/innovation/ai-and-data-science}{AT\&T Chief Data Office},
    \mkgsma \usenixhref{https://www.gsma.com/}{GSMA},
    \mkrelai \usenixhref{https://www.relational.ai/}{RelationalAI},
    \mkkhalifa \usenixhref{https://www.ku.ac.ae/}{Khalifa University},
    \mkkleeds \usenixhref{https://www.leeds.ac.uk/}{University of Leeds},\\
     \mkkyale \usenixhref{https://www.yale.edu/}{Yale University},
    \mkutdallas \usenixhref{https://utdallas.edu/}{The University of Texas at Dallas}
}

\renewcommand{\shortauthors}{OTel Team}

\begin{abstract}
Frontier AI models have advanced rapidly, but they still struggle with telecom-specific tasks. We present Open Telco (\texttt{OTel}), an open telecom AI resource with derived datasets for retrieval, reranking, instruction tuning, and safety/abstention, plus 30 full-parameter post-trained baselines across embedding, reranking, and language models.
The community has already engaged substantially with the resource: as of May 3, 2026, the released models have been downloaded over 16 million times, and the project has received 157+ pieces of media coverage worldwide. Building on prior open telecom datasets and benchmarks, \texttt{OTel} provides documented telecom data sources, held-out evaluation partitions, trained embedding models, rerankers, context-grounded LLMs, and safety/abstention data in one unified resource. %Each baseline starts from an open-weight model, is post-trained on \texttt{OTel}-derived data using an open training recipe, and is evaluated on held-out \texttt{OTel} evaluation partitions. 
\texttt{OTel} post-training improves performance across all three model families: embedding retrieval reaches 93.5\% NDCG@10, reranking reaches 0.952 MRR@10, and language-model correctness reaches 88.2\%. We release \texttt{OTel} as a reproducible starting point and invite the community to expand the data, improve embedding and reranking models, and build stronger context-grounded telecom LLMs.
\end{abstract}

\maketitle
\makeatletter \gdef\@ACM@checkaffil{} \makeatother

\renewcommand{\thefootnote}{\fnsymbol{footnote}}
\footnotetext[1]{\rm \mkletter \rm Corresponding authors: farbod.tavakkoli@att.com, lpowell@gsma.com, imtiaz.karim@utdallas.edu}

\section{Introduction}

Telecommunications infrastructure underpins nearly every aspect of modern digital life, yet it remains one of the most technically demanding domains for AI systems to navigate. The standards that govern it, 3GPP specifications running to hundreds of thousands of pages, O-RAN alliance documents, GSMA permanent reference documents, and IETF RFCs are dense, highly interdependent, and continuously
revised. A single specification can span dozens of releases; a single question about network behavior may require synthesizing content across multiple working groups and document versions. For AI systems to be genuinely useful in this environment, general-purpose language modeling is not sufficient. What is required is domain-specific training data at scale, retrieval pipelines tuned to the structure and vocabulary of telecom corpora, and evaluation benchmarks that reflect the kinds of questions practitioners actually ask.

The academic and industry research community has recognized this need and responded with a growing ecosystem of telecom-specific benchmarks. 
This research has focused on curating public 5G datasets for LLMs to utilize~\cite{karim-etal-2023-spec5g}, evaluating broad telecom knowledge and standards understanding across thousands of questions drawn from 3GPP and related sources~\cite{teleqna} and domain-specific mathematical reasoning~\cite{telemath}, focused specifically on O-RAN~\cite{oranbench}, targeting 5G root cause analysis in operational log data~\cite{telelogs}. The GSMA Open Telco AI Leaderboard~\cite{gsma_telco_leaderboard} brings these fragmented efforts together into a shared evaluation interface, making it possible to compare models across multiple telecom tasks in a single view. These benchmarks represent a significant collective achievement: they have established, with increasing precision, what telecom AI systems should be able to do.

What remains missing, however, is equally important: a unified open resource for \emph{building} the systems that can meet those benchmarks. Evaluation without training infrastructure creates an asymmetry. Practitioners who want to develop telecom-capable retrieval, reranking, or generation models must assemble their own training data from heterogeneous sources, design their own cleaning pipelines,
implement their own evaluation setups, and release their own baselines, all independently, without a shared foundation. This makes progress slow, comparisons unreliable, and reproducibility difficult. Domain AI ecosystems in medicine, law, and finance have shown that this asymmetry can be resolved by releasing not just benchmarks but also aligned training resources and reproducible baselines that
the community can build on directly.

\looseness-1
We address this gap with \textbf{Open Telco (\texttt{OTel})\footnote{\url{https://huggingface.co/farbodtavakkoli}}}, an open telecom AI resource that covers the full Retrieval-Augmented Generation (RAG) pipeline from data collection through model release. \texttt{OTel} is the product of a large-scale collaborative effort involving more than 100 domain experts from industry and academia. It releases four derived
datasets: \texttt{OTel}\texttt{-Embedding}, \texttt{-Reranker}, \texttt{-LLM}, and \texttt{-Safety} built from roughly 1.1 million raw training points spanning 3GPP specifications, GSMA documents, O-RAN alliance materials, RFCs, industry whitepapers, and telecom academic papers, and cleaned to 326,767 high-confidence examples through a rigorous multi-stage quality pipeline.
Alongside the data, \texttt{OTel} releases 30 full-parameter post-trained baselines covering embedding models, rerankers, and language models of varying sizes, each trained on the matching \texttt{OTel} dataset under a shared reproducible recipe and
evaluated on held-out \texttt{OTel} partitions.

The community response has been immediate and substantial. As of May 2026, the released \texttt{OTel} models have been downloaded over \textbf{16 million times} and the project has received \textbf{157+ pieces of media coverage} worldwide across telecom trade press, general technology outlets, and regional business media. This
level of engagement, within months of release, reflects the scale of unmet demand for open telecom AI infrastructure. Baseline results further validate the approach: \texttt{OTel} fine-tuning improves performance consistently across all three model families,
with embedding retrieval reaching \textbf{93.5\% NDCG@10}, reranking reaching \textbf{0.952 MRR@10}, and language-model correctness reaching \textbf{88.2\%}. We release \texttt{OTel} not as a finished product but as a reproducible starting point, a foundation the community can expand with new data, stronger models, multilingual coverage, and richer evaluation.

\section{Technical Contribution}

\texttt{OTel}'s technical contribution spans three interconnected layers: a curated dataset family derived from heterogeneous telecom sources, a suite of full-parameter post-trained baselines covering the RAG pipeline, and a reproducible evaluation protocol that ties the two together. Each layer is designed to be independently useful and collectively coherent.

\noindent \textbf{Dataset collection and curation.}
The \emph{\texttt{OTel} source corpus} encompasses six categories of publicly available telecom documents: GSMA permanent reference documents, 3GPP specifications, O-RAN alliance specifications across multiple working groups, the IETF RFC series, topical telecom material including eSIM and roaming documentation, and industry whitepapers and telecom academic papers. Over 100 domain experts contributed
approximately 1.1 million raw training points, with Yale University providing roughly 680K question-answer-source triples derived from arXiv telecom papers, 3GPP standards, telecom Wikipedia articles, and Common Crawl telecom pages; GSMA contributing 158K examples from PRDs and the Discover portal; NetoAI providing 100K RFC-derived examples; and Khalifa University, the University of Leeds, and UT Dallas together contributing over 160K O-RAN and whitepaper examples.

The incoming data presented two distinct structural challenges. The Yale contribution consisted of question-answer-source triples without explicit retrieval passages, where source documents can span hundreds of thousands of tokens. Converting these into retrieval-ready supervision required a dedicated six-stage enrichment pipeline: (i) grouping QA pairs by source document and joining them to the full document text; (ii) creating candidate passages via sliding window and semantic chunking; (iii) mining hard negatives both
within and across source documents, followed by reranker rescoring to improve negative quality; (iv) applying fact-grounded claim verification by decomposing reference answers into atomic claims and checking each against the candidate context; (v) selecting the minimal sufficient context through greedy minimization; lastly (vi) formatting the verified examples into contrastive training structures suitable for Multiple Negatives Ranking Loss and related retrieval objectives.

The need for aggressive cleaning was confirmed empirically. We trained separate full-parameter embedding models on independent data shards and evaluated each on a common held-out set. Performance varied dramatically, from 71.1\% to 3.9\% Acc@1 across shards, revealing substantial noise in the raw corpus. The final cleaning pipeline applied four sequential filters heuristic filtering to remove malformed and low-information examples, reranker-based semantic filtering to remove query-passage pairs with low semantic alignment, embedding-based semantic filtering to remove near-duplicates and low-coherence examples, and global deduplication. For the Yale subset, this reduced 680K examples to 220,334. For the remaining contributors, 420K examples were reduced to 106,433. The final
retained corpus contains \textbf{326,767} high-confidence training points.

These are released in four task-specific formats ordered by the RAG workflow. \texttt{OTel-Embedding} provides anchor, positive, and up to five hard-negative passages per example for contrastive retrieval training. \texttt{OTel-Reranker} provides query-passage pairs with binary relevance labels
for cross-encoder training. \texttt{OTel-LLM} provides context grounded instruction-tuning prompts with reference completions and abstention flags indicating whether the retrieved context
is sufficient to answer. \texttt{OTel-Safety} extends \texttt{OTel-LLM} with examples specifically designed
for abstention tuning, where retrieved context is off-topic or insufficient and the correct model behavior is to decline rather than generate. All four datasets are released under Apache-2.0 as derived QA pairs; raw source documents are not redistributed, and release documentation describes provenance, intended use, and source-specific constraints at the asset level.

\noindent \textbf{Post-trained baselines.}
\texttt{OTel} releases 30 full-parameter post-trained baselines: 10 embedding models ranging from 22M to 8B parameters, 3 rerankers ranging from 0.6B to 8B parameters, and 17 language models ranging from 270M to 32B parameters. Each baseline starts from a publicly available open-weight checkpoint including models from the Gemma~3, Qwen~3, OLMo~3, Mistral~3, LFM2, and RNJ-1 families and is full-parameter post-trained on the matching \texttt{OTel} dataset. %The shared training recipe uses AdamW in 8-bit form with cosine decay and warmup, BF16 precision, Flash Attention~2, gradient checkpointing, and Fully Sharded Data Parallel training. LLMs and embedding models train for three epochs; rerankers train for two. 
Training was distributed across AMD MI300X, MI325X, and MI355X
GPUs alongside NVIDIA A100 and H100 nodes, with large runs reaching 94.2\% GPU
utilization across 256 AMD MI325X GPUs.

\begin{figure}[t]
\centering
\includegraphics[width=0.45\textwidth]{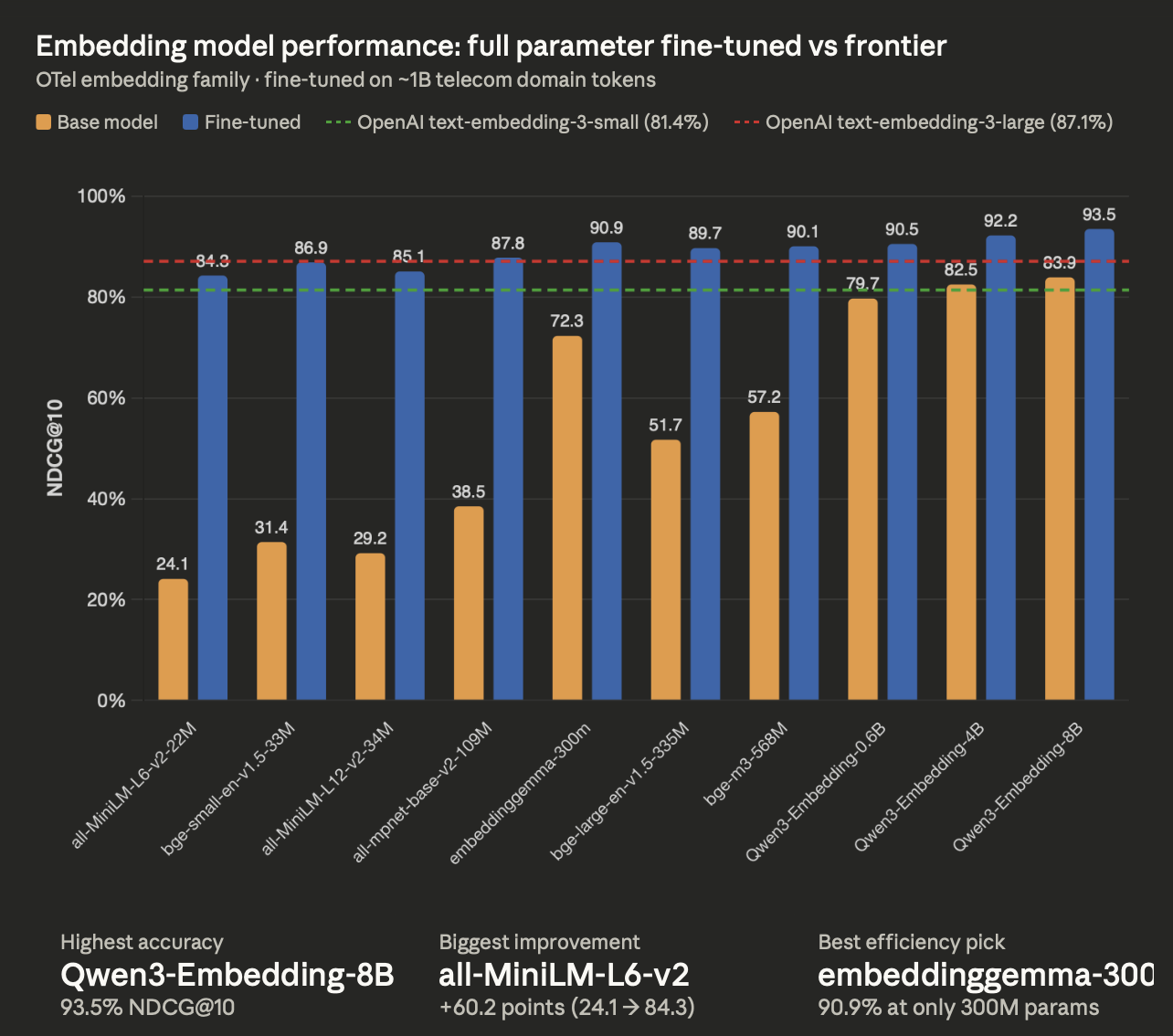}
\caption{Fine-tuning improvement of embedding retrieval} %; larger LLMs generally raise the correctness upper envelope.}
%\vspace{-0.3cm}
\label{fig:1}
\end{figure}

\begin{figure}[h]
\centering
\includegraphics[width=0.45\textwidth]{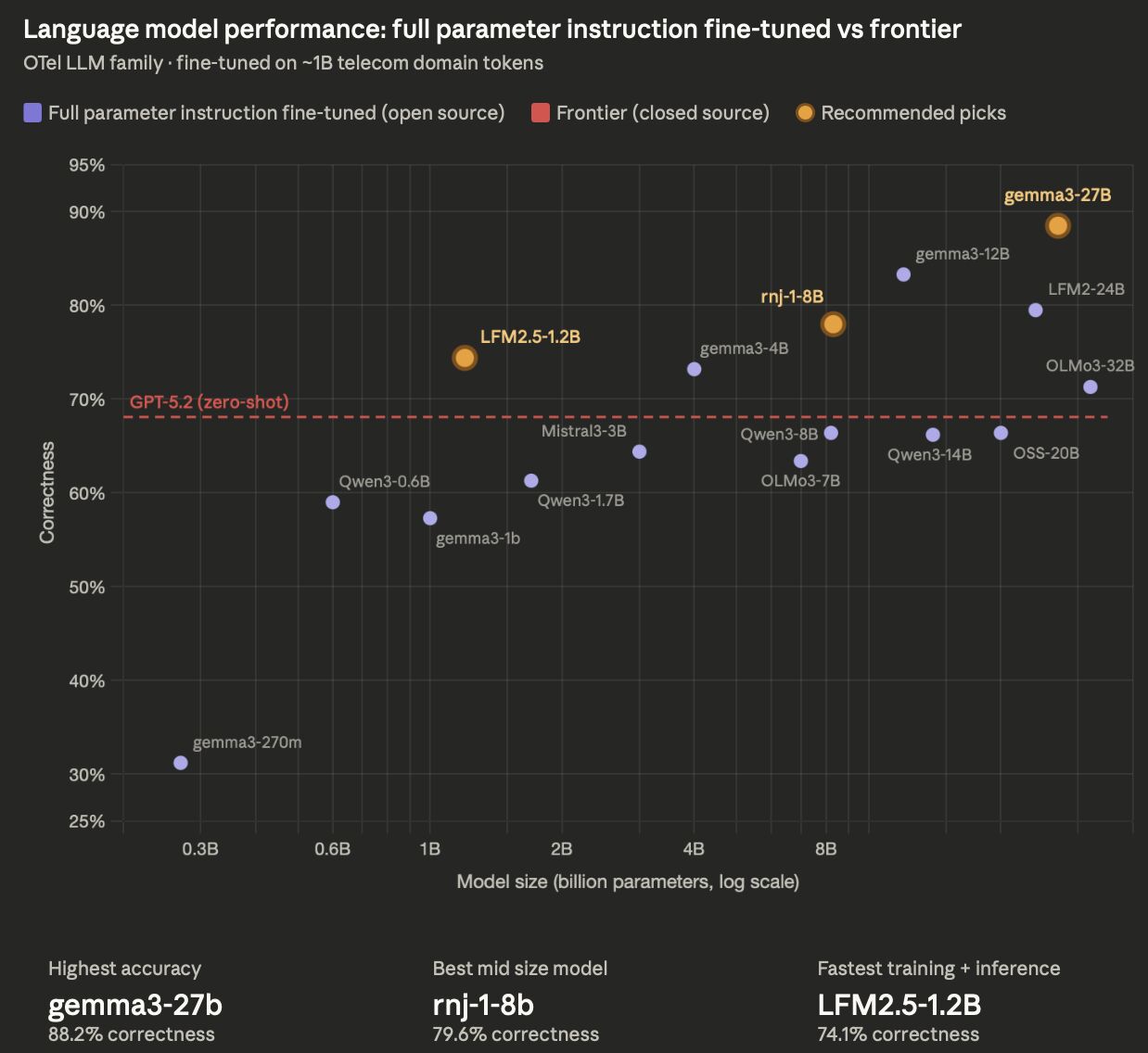}
\caption{Fine-tuning improvement of language model}
%\vspace{-0.6cm}
\label{fig:2}
\end{figure}

\texttt{OTel} fine-tuning improves performance consistently across all three model families and all parameter scales. For embedding models, NDCG@10 improves by +9.6 to +60.2 percentage points over the respective base models (shown in Figure~\ref{fig:1}). %Even the smallest
%released embedding model (22M parameters, based on all-MiniLM-L6-v2) reaches \textbf{84.3\%} \textbf{NDCG@10} after \texttt{OTel} fine-tuning, up from 24.1\% without it. The largest (8B parameters, based on Qwen3-Embedding-8B) achieves \textbf{93.5\%} \textbf{NDCG} \textbf{@10}. 
For rerankers, MRR@10 improves by +0.535 to +0.598 absolute across all three sizes, with all models exceeding 0.944 MRR@10 after fine-tuning, compared to 0.35--0.42 for the base models. For language models, LLM-as-judge correctness improves by +3.7 to +10.0 percentage points (shown in Figure~\ref{fig:2}). The largest model,
\texttt{OTel}\texttt{-LLM-27B-IT}, reaches \textbf{88.2\% correctness};
\texttt{OTel} \texttt{-LLM-8.3B-IT} is the strongest mid-size baseline at \textbf{79.6\%}, outperforming other models in its weight class by a substantial margin; \texttt{OTel-LLM-1.2B-IT} provides a competitive low-latency option at \textbf{74.4\%}. While larger models generally raise the upper envelope, the
results also show that architecture and training family matter independently of scale: \texttt{OTel-LLM-8.3B-IT} outperforms several models with two to four times as many parameters.

\noindent \textbf{Evaluation protocol.}
Held-out evaluation partitions follow a 90/10 train/eval split for LLMs and embeddings and a 95/5 split for rerankers, all with seed 42. LLMs are evaluated by GPT-4o mini LLM-as-judge correctness on context-grounded answer generation:
models receive retrieved telecom context and are judged on whether their answer is correct relative to that context and a reference answer. This scoping is intentional the \texttt{OTel} LLMs are designed for RAG, not unrestricted open-ended
QA, and results should be interpreted accordingly. Embeddings are evaluated by NDCG@10, which captures ranking quality across the top retrieved passages. Rerankers are evaluated by MRR@10, which captures how quickly the first truly relevant passage is promoted. All results include standard errors computed via
bootstrap resampling ($n{=}10$).

Collaborator-led retrieval stress tests from UT Dallas and the University of Leeds provide additional diagnostic evidence beyond the aggregate held-out splits. These tests evaluated O-RAN retrieval on fixed-chunk question pools and
examined retrieval behavior across 3GPP, GSMA PRD, O-RAN, RFC, whitepaper, and academic paper subdomains. The results are not reported as formal benchmark tables because protocols were not yet standardized, but they surface a consistent pattern: O-RAN retrieval is comparatively strong across models, while
academic paper and GSMA PRD examples remain weaker areas, suggesting that targeted curation and harder negatives in these subdomains should be a priority for future \texttt{OTel} releases.

\section{Broader Implications}

The significance of \texttt{OTel} extends well beyond its benchmark numbers. Telecom is among the most consequential and least AI-accessible technical domains: its standards are vast, its terminology is specialized, its documents are
interdependent, and the cost of errors in deployment, such as misconfigured networks, incorrect protocol interpretation, and unsafe automation, can be severe. \texttt{OTel} represents a systematic effort to make this domain accessible to the broader AI research community, and its design choices reflect lessons from analogous efforts in other high-stakes fields.

\noindent \textbf{Democratizing telecom AI research.}
By releasing aligned training data, post-trained baselines, and a reproducible evaluation protocol in a single unified resource, \texttt{OTel} removes the infrastructure barrier that has historically limited telecom AI research to organizations with
the resources to assemble all three independently. A research group that previously needed to negotiate data access, design a cleaning pipeline, train models from scratch, and build an evaluation harness can now start from \texttt{OTel}'s baselines and focus their effort on the specific improvement they want to make. This mirrors the role that shared resources have played in accelerating progress in medical AI following MedQA~\cite{medqa}, legal AI following LegalBench~\cite{legalbench}, and financial AI following FinBen~\cite{finben}. In each of those domains, the
availability of shared training data and reproducible baselines compressed years of scattered effort into a coherent research agenda. \texttt{OTel} is designed to play the
same role for telecom.

\noindent \textbf{Industry uptake and real-world validation.}
The scale of community engagement 16 million model downloads and 157+ media mentions within months of release provides early evidence that \texttt{OTel} is addressing a genuine and widespread need. This is not merely academic interest:
the coverage spans telecom trade press, general technology outlets, and regional business media, reflecting demand from practitioners across the industry who need domain-specific AI tools and have not previously had access to open, reproducible
starting points. The multi-organizational authorship of \texttt{OTel} spanning a major US carrier, a global industry consortium, academic institutions on four continents, and hardware and software infrastructure partners also reflects a
level of cross-sector coordination that is rare in AI research and that positions \texttt{OTel} as a genuinely community-owned resource rather than a single organization's
release.

\noindent \textbf{Responsible deployment and safety.}
Responsible deployment is central to \texttt{OTel}'s design rather than an afterthought. The context-grounded RAG framing is itself a safety mechanism: by training models
to answer from retrieved context rather than from parametric memory alone, \texttt{OTel} reduces the risk of fluent but incorrect outputs on technical telecom questions.
The \texttt{OTel-Safety} dataset and its associated abstention-tuned model variants extend this further, training models to recognize when retrieved context is insufficient or off-topic and to decline rather than confabulate. Model cards, dataset cards, Croissant 1.1 metadata with full Responsible AI field coverage, and asset-level provenance documentation are included for all 30 baselines and four datasets. %All derived data is released under Apache-2.0, and source-specific licensing constraints are documented at the asset level rather than obscured behind a single license claim.

\looseness-1
\noindent \textbf{Limitations and future directions.}
\texttt{OTel} is an open starting point, not a finished product, and its current limitations define a concrete research agenda. The resource is English-only and text-centric, which excludes significant portions of the global telecom community and omits the diagrams, tables, and structured data that are central to
many telecom documents. The main-paper evaluation relies on held-out splits from the \texttt{OTel} corpus itself rather than fully independent external benchmarks, which limits the ability to assess cross-domain generalization. Collaborator
stress tests reveal meaningful subdomain variation that aggregate scores obscure. Finally, the current release does not include per-subdomain reporting as a first-class evaluation output, making it difficult to track improvements in specific areas over time.

Future \texttt{OTel} releases should address these gaps systematically: expanding to multilingual and multimodal data, improving academic-paper and GSMA PRD subsets with better negatives and larger evaluation pools, establishing per-subdomain
metrics as a standard part of the evaluation protocol, and validating on external benchmarks beyond \texttt{OTel}'s own held-out partitions. The resource is architected to support this kind of incremental community-driven growth new data can be
contributed through the same enrichment and cleaning pipeline, new models can follow the same training recipe and naming convention, and new evaluation subsets can be added to the held-out partitions without breaking existing comparisons. We
invite the community to build on \texttt{OTel}, improve what it provides, and push telecom AI toward the reliability and coverage that the domain demands.

\begin{acks}
Apart from the core \texttt{OTel} team, we also would like to acknwoledge the support of Kostikey Mustakas from AT\&T Chief Data Office; Antti-Ville Suni, Andrey Ivannikov, Andy Allred, Mark van Heeswijk, Alexander Finn, Kumaran Siva from AMD; Enrique Molero from GSMA; Molham Aref, Nikolaos Vasiloglou
from RelationalAI; David Kanter from MLCommons; Rick Lievano, Ven Kumar, Inayat Wali, Thomas Steagall from Microsoft; Mashroor Hasan Bhuiyan from University of Texas at Dallas; Mirza Masfiqur Rahman and Faik Kerem Ors from Purdue University Merouane Debbah, Esraa Fahmy, Bohao Wang
from Khalifa University; Zeinab Nezami, Shehr Bano
from University of Leeds; 
Leandros Tassiulas and Rex Ying from Yale University,
Matt Upson and Nick Sorros from 
Mantis NLP;  Vignesh Ethiraj, Ashwath David from 
NetoAI. 
\end{acks}

\bibliographystyle{ACM-Reference-Format}
\bibliography{sample-base}

\end{document}